%% file: arxiv.tex
\documentclass[10pt,twocolumn,letterpaper]{article}

\usepackage{iccv}
\usepackage{times}
\usepackage{epsfig}
\usepackage{graphicx}
\usepackage{amsmath}
\usepackage{amssymb}

\usepackage[font=scriptsize,labelfont=bf]{caption}

\usepackage[belowskip=-5pt,aboveskip=3pt]{caption} 

\usepackage{multirow}
\usepackage{tabularx}
\usepackage{dsfont}
\usepackage{url}
\usepackage{color}
\usepackage{subcaption}
\usepackage{amsthm}
\usepackage[export]{adjustbox}
\usepackage{comment}

\usepackage[utf8]{inputenc} 
\usepackage[T1]{fontenc}    
\usepackage{url}            
\usepackage{booktabs}       
\usepackage{amsfonts}       
\usepackage{nicefrac}       
\usepackage{microtype}      
\usepackage{enumitem}

\definecolor{dg}{rgb}{0,0.694,0.298}
\definecolor{purple}{rgb}{0.4,0.176,0.569}

\usepackage{pifont}
\definecolor{royalblue}{RGB}{65,105,225} 
\usepackage[pagebackref=true,breaklinks=true,colorlinks,citecolor=royalblue,bookmarks=false]{hyperref}

\def\iccvPaperID{1402} 

\ificcvfinal\fi
\iccvfinalcopy

\begin{document}

\title{CarveNet: Carving Point-Block for Complex 3D Shape Completion}

\author{Qing Guo\textsuperscript{1}$^{*}$, 
\,Zhijie Wang\textsuperscript{2}\thanks{Qing Guo and Zhijie Wang are co-first authors and contribute equally.},
\,\,Felix Juefei-Xu\textsuperscript{3},
Di Lin\textsuperscript{4},
Lei Ma\textsuperscript{2,5},
Wei Feng\textsuperscript{4},
Yang Liu\textsuperscript{1}\\~\\
\textsuperscript{1}\,Nanyang Technological University, Singapore, ~~
\textsuperscript{2}\,University of Alberta, Canada \\
\textsuperscript{3}\,Alibaba Group, USA, ~~
\textsuperscript{4}\,Tianjin University, China \\
\textsuperscript{5}\,Alberta Machine Intelligence Institute, Canada
}
\maketitle

\input{S0_abstract}
\input{S1_intro}
\input{S2_related}
\input{S3_intuitive}
\input{S4_method}

\input{S5_experiment}
\input{S6_conclusion}

{\small
\bibliographystyle{ieee_fullname}
\bibliography{ref}
}

\end{document}

%% file: S0_abstract.tex
\begin{abstract}

3D point cloud completion is very challenging, because it heavily relies on the accurate understanding of the complex 3D shapes (\eg, high-curvature, concave/convex, and hollowed-out 3D shapes) and the unknown~\&~diverse patterns of the partially available point clouds. In this paper, we propose a novel solution, \ie, \emph{Point-block Carving} (PC), for completing the complex 3D point cloud completion. Given the partial point cloud as the guidance, we carve a 3D block that contains the uniformly-distributed 3D points, yielding the entire point cloud. To achieve PC, we propose a new network architecture, \ie, \emph{CarveNet}. This network conducts the exclusive convolution on each point of the block, where the convolutional kernels are trained on the 3D shape data. CarveNet determines which point should be carved, for effectively recovering the details of the complete shapes. Furthermore, we propose a sensor-aware method for data augmentation, \ie, \emph{SensorAug}, for training CarveNet on richer patterns of partial point clouds, thus enhancing the completion power of the network. The extensive evaluations on the ShapeNet and KITTI datasets demonstrate the generality of our approach on the partial point clouds with diverse patterns. On these datasets, CarveNet successfully outperforms the state-of-the-art methods.

\end{abstract}

%% file: S1_intro.tex
\section{Introduction}\label{sec:intro}

3D point cloud has been regarded as one of the best representations of the 3D object. It is widely adopted in an array of robotic-relevant applications, such as the simultaneous localization and mapping (SLAM) \cite{zhang2014loam}, place recognition \cite{angelina2018pointnetvlad}, object detection \cite{shi2020pv}, LiDAR processing for autonomous driving \cite{gao2018object}, \etc. To achieve the 3D point cloud of the entire object reasonably well, it is necessary for using many range finders to capture the geometric data from different views, and conducting the accurate registration among the captured data. However, the expensive, heavy and energy-consuming finders are restricted to only a few affordable applications in practice. Yet, the sparse 3D points can often lose the geometric and semantic information, thus giving rise to the performance degradation of the robotic system.

\begin{figure}
\centering
\includegraphics[width=\linewidth]{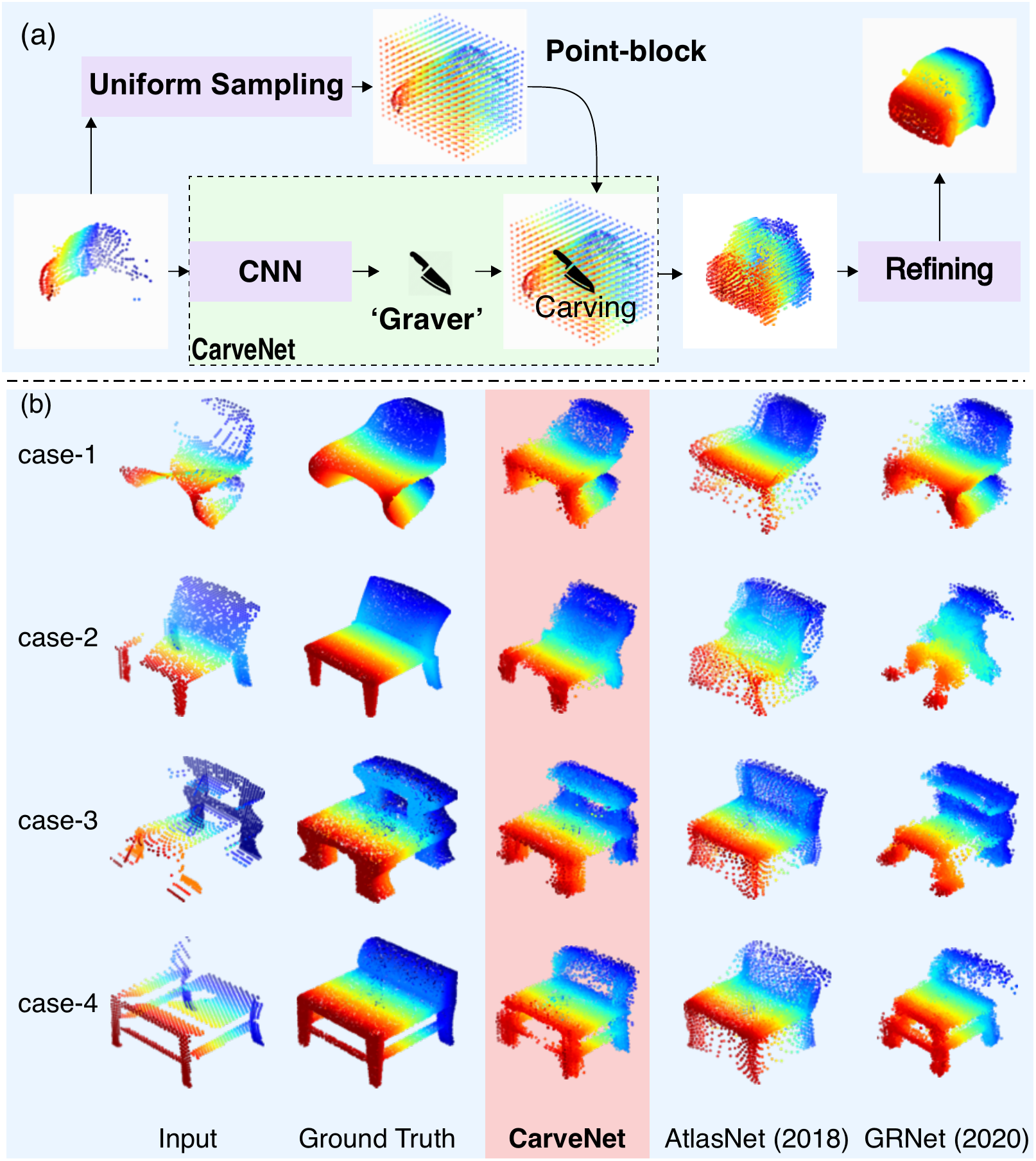}
\caption{(a) The intuitive idea and pipeline of our point-block carving with CarveNet. (b) The point cloud completion by AtlasNet \cite{groueix2018papier}, GRNet \cite{xie2020grnet}, and CarveNet on four cases with three complex shapes: high-curvature (case-1 and case-2), concave/convex (case-2 and case-3) and hollowed-out (case-3 and case-4).}\vspace{-10pt}
\label{fig:idea}
\end{figure}

The emergency of the point cloud completion methods alleviates the negative effect of the imperfect point cloud on the downstream applications. The current methods still face challenges when dealing with the complex object shapes, such as the shapes (see Fig.~\ref{fig:idea}) with high curvatures (\eg, $1^{st}$ case), concave/convex structures (\eg, $2^{nd}$ and $3^{rd}$ cases), and hollowed-out structures (\eg, $3^{rd}$ and $4^{th}$ cases). The methods~\cite{yang2018foldingnet,yuan2018pcn,groueix2018papier,tchapmi2019topnet, liu2020morphing, wen2020point, xie2020grnet} that focus on some kinds of complex shapes may lose the generality power for handling the completion of other shapes. For example, the state-of-the-art method GRNet \cite{xie2020grnet} loses some key structures in the $2^{nd}$ and $4^{th}$ cases.

Similarly, although the advanced methods such as AtlasNet \cite{groueix2018papier} employ the parameterized shape for flexible completion, the results lack the important visual details (\eg, the concave and hollowed-out structures in $2^{nd}$ and $3^{rd}$ cases). Moreover, given the similar (or even the identical) object shapes, the partial point clouds captured by different sensors may be variant. This is because the intrinsic performance of the sensors heavily affects the visual patterns of the partial point clouds. The variant information can easily mislead the completion methods, leading to the inconsistent completion results. This post-pressing concerns especially for real-world applications and calls for better methods to reduce the impact of the variant sensors on the point cloud completion.

In this paper, we propose a brand-new \emph{point-block carving} (PC). Given the partial point cloud as the input data, our approach leverages a set of "gravers" to carve the point-block to approximate the underlying object shape as similar as possible (see Fig.~\ref{fig:idea}). Intuitively, our carving process can be understood as the carving for a statue, where the partial shape of the statue is provided as a hint. Here, our carving process is done on the 3D block, where we add in a set of uniformly-distributed 3D points at the beginning. We also register the partial point cloud is the 3D block. Next, we use the \emph{CarveNet} to generate the graver for processing the 3D block and recovering the object shape. This is done by using the graver to remove the redundant 3D point. Specifically, the graver is defined by the point-wise convolution. The convolutional kernels are learned from the prior knowledge of the object that capture the geometric and semantic relationships between the existing and the missing 3D points in the 3D block. The graver propagates the useful prior information from the partially-given points to the add-in uniform points, whose present/absent statuses are predicted for recovering the lost part of the point cloud. We only process the add-in 3D points for effectively reducing the carving complexity. Note that the features, which are produced by different convolutional layers of CarveNet, contain rich semantic information of the objects. We use these features to regress new 3D points. These 3D points are added to the block to form denser completion result.

Moreover, we propose a new \emph{SensorAug} for augmenting the training data. SensorAug works with a similarity loss. Given a complete point cloud, we assume that a LiDAR sensor is randomly put for observing the object, leading to some invisible points. These invisible points are removed from the complete cloud, producing the partial cloud. Based on the same complete point cloud, the similarity loss of SensorAug involves more diversity of the partial point clouds. It allows CarveNet to be trained on more diverse data, for enhancing the completion power on the complex shapes.

We evaluate our method on the ShapeNet and KITTI datasets. With CarveNet, we successfully improve the completion performance on different benchmarks, even on the challenging cases where the critical points are unavailable for predicting the complex object shapes. Our method helps to achieve better results than the state-of-the-art methods. Our contribution is manifold:

\begin{itemize}[nosep,nolistsep,leftmargin=*] 
    \item We promote a novel paradigm, based on the carving of the point-block, for point cloud completion.
    \item We originally propose CarveNet for point-block carving, which facilitates better completion of the complex shapes.
    \item We propose SensorAug for data augmentation. SensorAug enhances the generality power of the completion model and helps the model to achieve the state-of-the-art results on the public benchmarks.
\end{itemize}
\begin{figure*}
\centering
\includegraphics[width=0.95\linewidth]{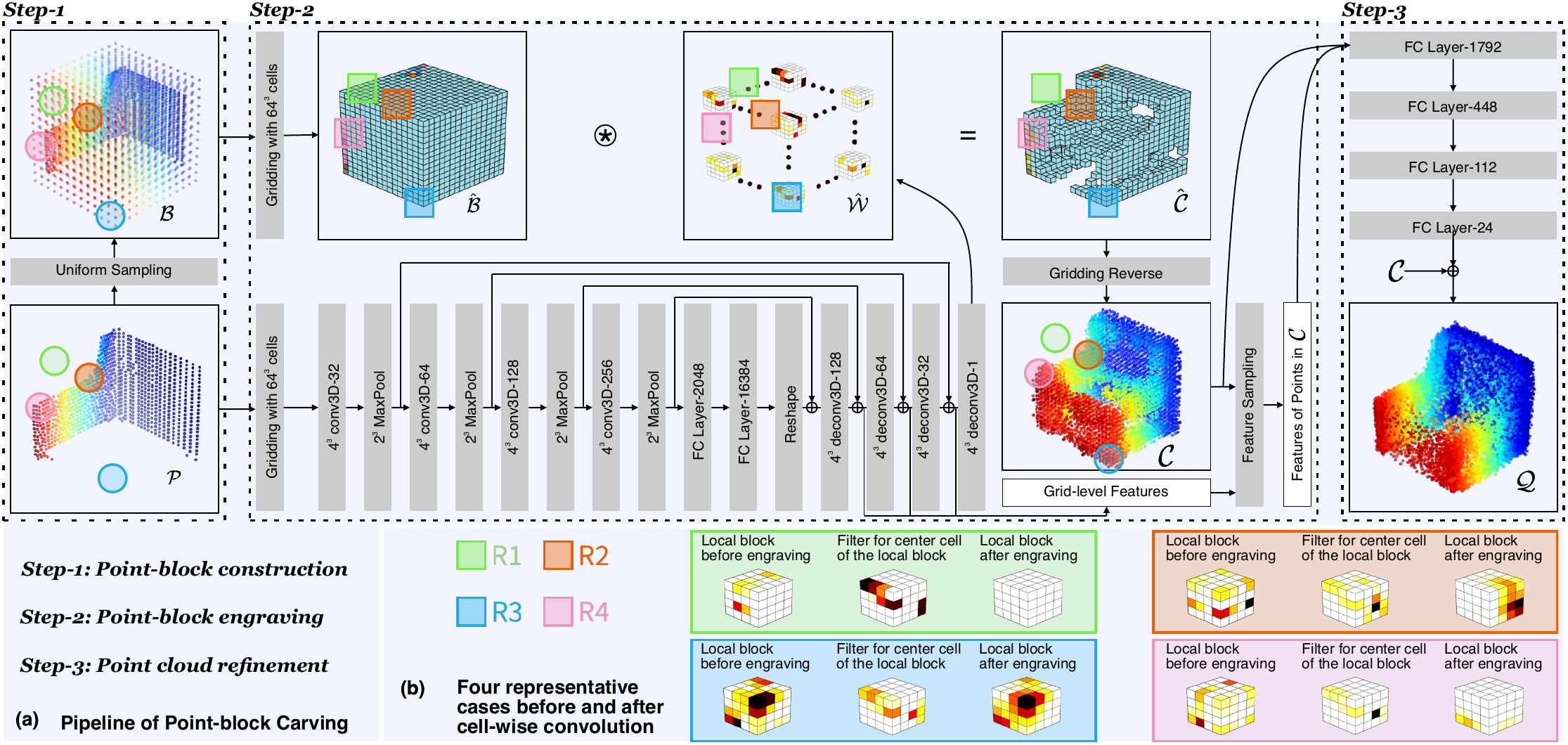}
   \caption{(a) The point-block carving achieved by CarveNet. (b) Four
   examples of the cell-wise convolution on different cells of the point-block.}
\label{fig:carvenet}\vspace{-10pt}
\end{figure*}

%% file: S2_related.tex
\section{Related Works}\label{sec:related}

We mainly discuss the deep-learning networks with different architectures for 3D point cloud completion. We also discuss different representations of the point cloud.

\noindent{\bf Network architectures.~~}
The deep network has been used for constructing many advanced point cloud completion methods. The \emph{folding-based} and \emph{MLP-based} network are two kind of the popular architectures.

The literature on the folding-based architecture is vast. FoldingNet \cite{yang2018foldingnet} folds 2D grid points twice into the target object's surface, with the guidance of the feature vectors extracted by PointNet \cite{qi2017pointnet}. But the complex object shape significantly increases the difficulty of folding the object, at the cost of expensive computation. In contrast, AtlasNet \cite{groueix2018papier} separates the object's surface into small patches. Each patch is folded individually. MSN \cite{liu2020morphing} employs the expansion loss to deal with the overlapping between different patches. PCN \cite{yuan2018pcn} is equipped with a two-stage completion. PCN uses MLPs to predict the coarse shape, which is processed by the deformation of the 2D grids for the denser completion result.

There have been many works based on the MLP-based architecture. TopNet \cite{tchapmi2019topnet} involves a tree-structure decoder, where MLP is used to connect the tree nodes. MLP can be used for producing the multi-scale features \cite{wen2020point,huang2020pf} to assist the completion. Several works \cite{huang2020pf, wang2020cascaded, wang2020point, sarmad2019rl} explore the generative adversarial networks (GAN) \cite{goodfellow2014generative} to construct MLPs to produce more realistic completion results.

In this paper, we carve the 3D block to remove the shape-irrelevant points and use CarveNet to learn from the diversity of the object shapes. Compared to the folding-based networks, CarveNet is better especially in terms of completing the complex 3D shapes, for yielding the completion results with richer details and less redundant points. CarveNet is equipped with MLP for refining the coarse completion results. In contrast to the current MLP-based architectures, we use different layers of MLPs, providing the semantic object information at different levels, to produce denser results.

\noindent{\bf Point cloud representations.~~}
Point cloud representations play a key role in the related tasks. A simple point cloud representation is achieved by directly voxelizing the 3D points. However, this voxelization results are sensitive to the quantization effects \cite{dai2017shape, han2017high, sharma2016vconv, stutz2018learning, le2018pointgrid}. PointNet learns point-wise features directly from the raw data of point clouds. PointNet++ \cite{qi2017pointnet++} group the points hierarchically to achieve the flexible receptive fields on the 3D space. SO-Net \cite{li2018so} builds a self-organizing map to represent the spatial relationships of the unordered points. PointCNN \cite{li2018pointcnn} learns the $\mathcal{X}$-transformation of the input points to a latent and ordered representation. KPConv \cite{thomas2019kpconv} uses the pseudo-grid convolution on the equally distributed spherical points. In the local feature aggregation method \cite{liu2020closer}, the position-pooling is used for extracting the local features efficiently.

In this work, we resort to the state-of-the-art representation, \ie, 3D grid-based intermediate representation \cite{xie2020grnet}. The point cloud can be easily registered to the regular 3D grid, without losing the key object structures. The 3D grid can be processed by CarveNet, and the output of CarveNet can be converted back to the point cloud, efficiently.

%% file: S3_intuitive.tex
\section{Discussion on the Point-Block Carving}

3D point cloud completion relies on the understanding of the object shape. Intuitively, the object shape captures the object-level semantic information and the point-level spatial relationship. The conventional encoder-decoder methods (\eg, AtlasNet \cite{groueix2018papier} and GRNet \cite{xie2020grnet}) take input as the partial point cloud. They map the points to the high-level features, which are merged to form the object-level semantic information, for determining the locations of the missing points. Actually, the objects, which even belong to the same category, likely have different structural details. It means that the spatial relationship between the 3D points are complex, hardly determined by the object-level information alone.

Our method for point-block carving takes the advantage of the both object-level information and point-level spatial relationship for completion. 
We construct the point-block containing uniformly distributed 3D points and the input partial point cloud.
We calculate the high-level features of the partial point cloud by the deep network.
Rather than using the high-level features to predict the locations of the missing points, we use these features to learn the kernels. These kernels are used for propagating the semantic information between the points and their neighbors in the point-block, thus also capturing the underlying spatial relationship between the points. With more complex spatial relationship available, the redundant points can be removed from the point-block, by respecting the details of the object shape.

%% file: S4_method.tex
\section{Point-block Carving for Shape Completion}
\label{sec:method}

We denote the partial point cloud as a set of 3D points, \ie, $\mathcal{P}=\{\mathbf{p}_i \in \mathbb{R}^{3} ~|~ i=1,...,|\mathcal{P}|\}$, where $|\mathcal{P}|$ is the number of 3D points. Based on $\mathcal{P}$, we use the point-block carving to compute a new set of points $\mathcal{Q}=\{\mathbf{p}_i \in \mathbb{R}^{3} ~|~ i= 1,...,|\mathcal{Q}|\}$ to represent the complete object. The set $\mathcal{Q}$ and the ground-truth set $\mathcal{G}=\{\mathbf{p}_i \in \mathbb{R}^{3} ~|~ i=1,...,|\mathcal{G}|\}$ should be as similar as possible. We construct CarveNet to achieve the carving process. In Fig.~\ref{fig:carvenet}, we schematically illustrate CarveNet, which consists of the point-block construction, the point-block engraving, and the point cloud refinement.

\noindent{\bf Point-block construction.}
First, we compute a 3D block that contains the partial point cloud $\mathcal{P}$. The range of the 3D block is denoted as $x\in[x_\text{min},x_\text{max}]$, $y\in[y_\text{min},y_\text{max}]$, and $z\in[z_\text{min},z_\text{max}]$, where $x_\text{min}$/$y_\text{min}$/$z_\text{min}$ (or $x_\text{max}$/$y_\text{max}$/$z_\text{max}$) represents the possible minimization (or maximization) x/y/z-coordinate in the complete point cloud $\mathcal{Q}$. In the practice, the range can be estimated through a 3D bounding box detection method \cite{xu2018pointfusion}. Next, we sample $N$ 3D points from the 3D space of the block, where the $N$ 3D points distribute uniformly. The sampled 3D points are used along with the partial point cloud $\mathcal{P}$ to form the point-block $\mathcal{B}=\{\mathbf{p}_i ~|~ i=1,...,|\mathcal{P}|+N\}$.

\begin{figure}[tb]
\centering
\includegraphics[width=\linewidth]{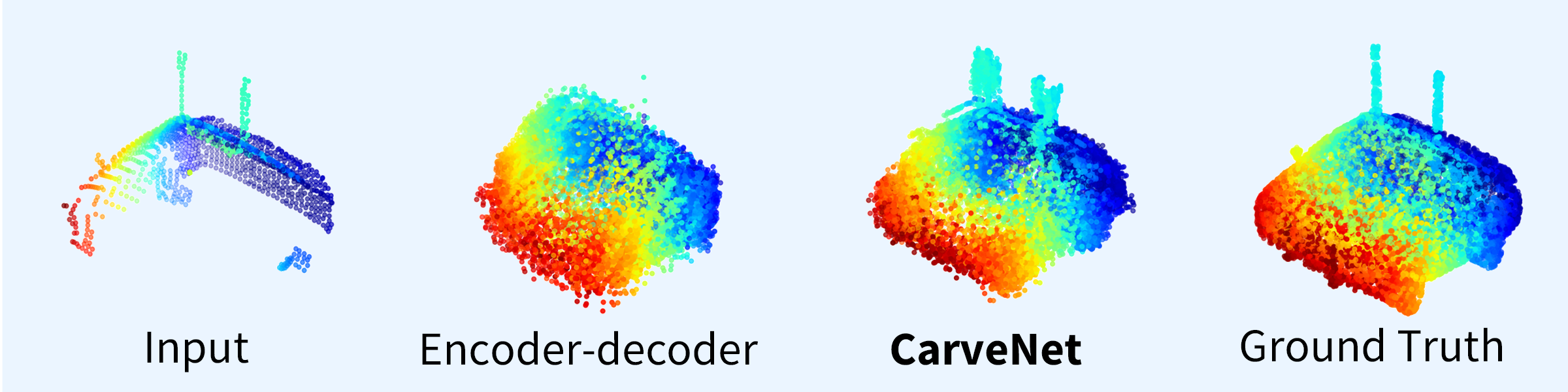}
\caption{The comparison between the completion results by using the state-of-the-art encoder-decoder method (GRNet) and our point-block engraving, where our method preserves better details of the object shape in the completion result.}
\label{fig:autoenco-carve}
\end{figure}

\noindent{\bf Point-block engraving.} We use the point-block $\mathcal{B}$ to compute the coarse completion $\mathcal{C}$. This is done by using CarveNet to process the 3D grid-based intermediate representation of $\mathcal{B}$, where the irrelevant points are removed. CarveNet considers the visual and spatial properties of each 3D point, learning the point-wise graver with unique parameters to manipulate the 3D point. Thus, compared to the current encoder-decoder network like GRNet \cite{xie2020grnet} that processes different 3D points by using the identical set of network parameters, CarveNet preserves better details during the carving process (see Fig.~\ref{fig:autoenco-carve}). The 3D grid-based intermediate representation enables the linear interpolation of the grid-level features for computing the deep features of the coarse points. It helps to refine the point cloud finally. We provide more details of CarveNet and the 3D grid-based intermediate representation of the coarse point cloud in Sec.~\ref{subsec:carvenet}.

\noindent{\bf Point cloud refinement.} We refine the coarse point cloud $\mathcal{C}$ for achieving the dense point cloud $\mathcal{Q}$. Here, we use CarveNet to extract a set of features for all points in the coarse point cloud. The feature of each 3D point is concatenated with the point's coordinate, which is passed to four fully-connected layers (with 1792, 2448, 112, 24 neurons in each layer, respectively).
We use the fully-connected layers to compute a set of point-wise offsets for updating the locations of the points in the coarse point cloud. The coarse point cloud and the updated counterpart are merged, for forming the dense point cloud $\mathcal{Q}$.

\subsection{CarveNet for Point-block Engraving} \label{subsec:carvenet}

\noindent{\bf Formulation.}
Given the point-block $\mathcal{B}$, we construct CarveNet to compute the point-wise gravers to remove the irrelevant points. More formally, we perform the exclusive, point-wise convolution on each 3D point in $\mathcal{B}$ as:
\begin{align}\label{eq:carve}
\mathcal{C}(\mathbf{p}_i)= (\mathcal{B}\circledast\mathcal{W})(\mathbf{p}_i)=\sum_{\mathbf{p}_j\in\mathcal{N}(\mathbf{p}_i)}w_i(\mathbf{p}_j-\mathbf{p}_i)f_j,
\end{align}
%
where $\mathcal{B}\circledast\mathcal{W}$ denotes the point-wise convolution. $(\mathcal{B}\circledast\mathcal{W})(\mathbf{p}_i)$ is the convolutional output for the point $\mathbf{p}_i$. $w_i$ defines the exclusive convolutional parameters for $\mathbf{p}_i$. $f_j$ is the feature of $\mathbf{p}_j$, and it is set to 1.

The present/absent status of the point $\mathbf{p}_i$ is determined by the graver, whose parameters are represented by $w_i$ in Eq.~\eqref{eq:carve}. To adjust the graver \wrt each 3D point dynamically, we resort to a 3D CNN to learn the graver from the partial point cloud $\mathcal{P}$. The 3D CNN outputs a set of graver parameters $\mathcal{W} = \{w_1,...,w_N\}$ for all of the $N$ sampled points in $\mathcal{B}$ as:
%
\begin{align}\label{eq:graver}
\mathcal{W}= \text{3DCNN}(\mathcal{P}).
\end{align}
%
We use the 3DCNN along with the 3D grid-based intermediate representation to implement the convolutional operations in Eq.~\eqref{eq:carve} and \eqref{eq:graver}. We remark that the alternatives, including KPConv \cite{thomas2019kpconv}, PointConv \cite{wu2019pointconv}, can be used in place of 3DCNN here. We compare different strategies of implementation in the experiment, in terms of the completion accuracy.

\noindent{\bf 3D grid-based intermediate representation.}
We choose the 3D grid-based intermediate representation \cite{xie2020grnet}. This representation is based on the neighboring interpolation. It effectively avoids any quantization, thus preserving the structural information of the object. Its advantages has been evidenced in the tasks \cite{qi2017pointnet,qi2017pointnet++,thomas2019kpconv,tchapmi2019topnet,xie2020grnet} where the point cloud needs to be processed efficiently.

In the intermediate representation, the \textit{gridding layer}, which uses the differentiable interpolation to map a 3D point cloud to the gridding representation. Conversely, the \textit{gridding reverse} maps a gridding to the 3D point cloud. The \textit{feature sampling} extracts the feature of the 3D point based on the grid-level features. We construct the 3D grid-based intermediate representation for the point cloud to regularize the unordered points, while explicitly preserving the structural and context information of these points. In Fig.~\ref{fig:carvenet}, we illustrate the construction of the intermediate representation.

We use two gridding layers, which compute the gridding results of the point-block $\mathcal{B}$ and the partial $\mathcal{P}$, respectively. The gridding results are denoted as $\hat{\mathcal{B}}\in\mathbb{R}^{H\times W\times M}$ and $\hat{\mathcal{P}}\in\mathbb{R}^{H\times W\times M}$. $H$, $W$, and $M$ denote the resolution of the grid. Here, we regard the grid as a set of $H\times W\times M$ cells.

Next, we process the $\mathbf{c}_i$ in $\hat{\mathcal{B}}$ as:
\begin{align}\label{eq:carve_grid}
\hat{\mathcal{C}}(\mathbf{c}_i)=(\hat{\mathcal{B}}\circledast\hat{\mathcal{W}})(\mathbf{c}_i)=\sum_{\mathbf{c}_j\in\mathcal{N}(\mathbf{c}_i)}\hat{w}_i(\mathbf{c}_j-\mathbf{c}_i)\hat{\mathcal{B}}(\mathbf{c}_j),
\end{align}
where $\hat{\mathcal{W}}\in\mathbb{R}^{H\times W\times M\times K^3}$ is a set of 3D cell-wise kernels. $K^3$ indicates the size of a 3D kernel. We use a pre-trained 3D UNet to process the grid $\hat{\mathcal{P}}$, achieving the $\hat{\mathcal{W}}$ as:
\begin{align}\label{eq:graver_grid}
\hat{\mathcal{W}}= \text{UNet}(\hat{\mathcal{P}}).
\end{align}
We illustrate 3D UNet in Fig.~\ref{fig:carvenet}. We represent the griding representation of the coarse point cloud as $\hat{\mathcal{C}}=\hat{\mathcal{B}}\circledast\hat{\mathcal{W}}$. In Fig.~\ref{fig:carvenet}~(b), we provide four examples of the cell-wise convolution on different cells in $\hat{\mathcal{B}}$.
We compute the gridding reverse~\cite{xie2020grnet} of $\hat{\mathcal{C}}$, achieving the final coarse point cloud $\mathcal{C}$.

We extract features from the $\text{UNet}(\hat{\mathcal{P}})$. The features are propagated to the points in $\mathcal{C}$, by resampling \wrt the coordinate relationship between $\mathcal{C}$ and $\hat{\mathcal{P}}$ (see the ``Feature Sampling" in Fig.~\ref{fig:carvenet}~(a)). These features are used to refine the coarse point cloud by MLP.

\begin{figure}[t]
\centering
\includegraphics[width=\linewidth]{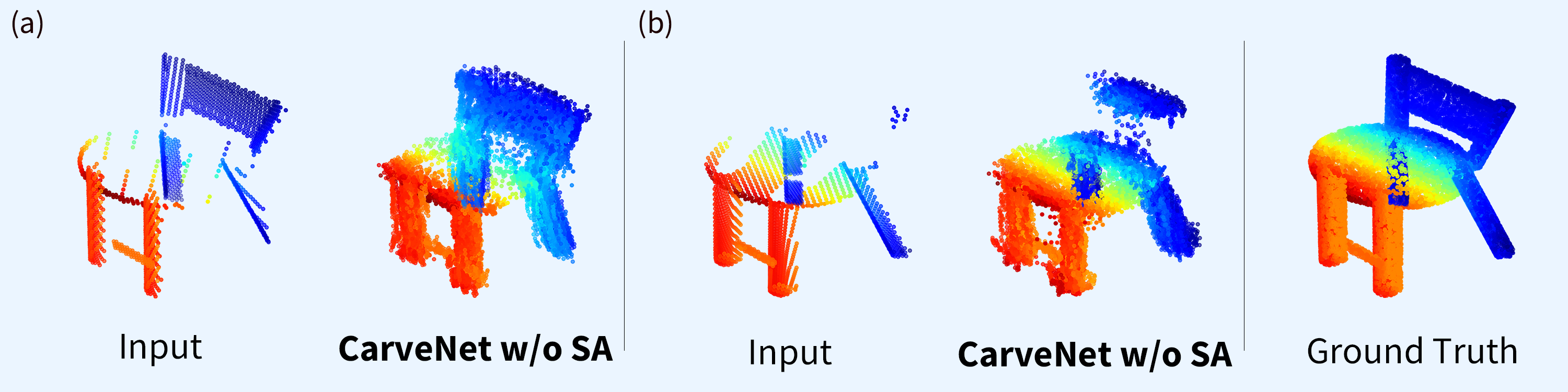}
\caption{(a) and (b) are different partial point clouds for the same object. The completion results of (a) and (b) with our method without the SensorAug (\ie, CarveNet~w/o~SA) are inconsistent.}
\label{fig:wodropmix}\vspace{-0.2in}
\end{figure}

\subsection{SensorAug for Training CarveNet}\label{subsec:dropmix}

We propose SensorAug for augmenting the training data. Rather than randomly removing 3D points from the complete point clouds, we move
a virtual sensor in the 3D space, letting the visible points to be the partial point could.

\begin{figure}[t]
\centering
\vspace{-0pt}
\includegraphics[width=\linewidth]{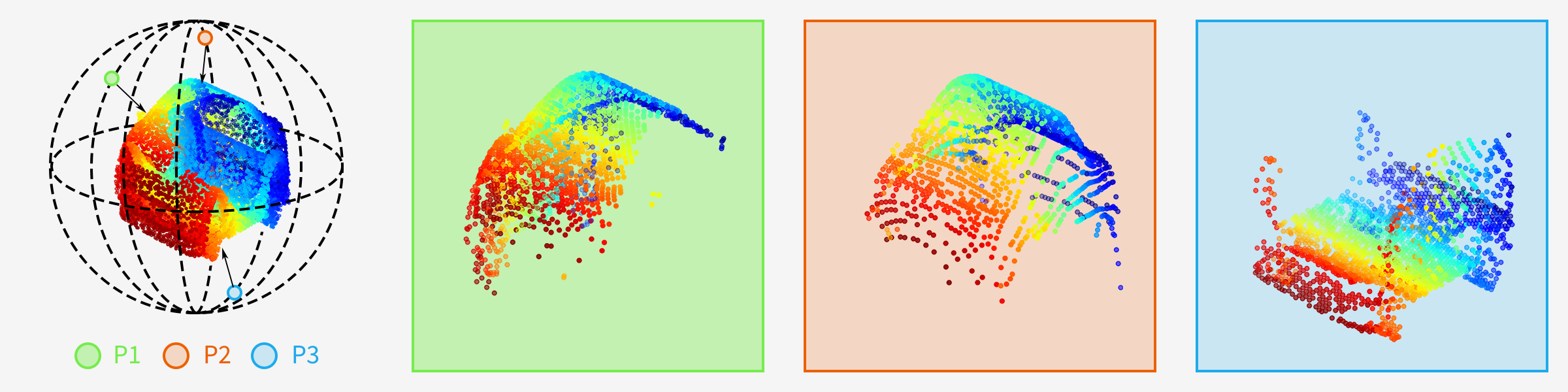}
\caption{The left-most subfigure illustrates the complete point cloud and the sensors at different locations. P1, P2, and P3 are different generated partial point clouds, which contain the visible points from different views.}
\label{fig:dropmix}\vspace{-0.2in}
\end{figure}

SensorAug works with a similarity loss function for supervising the training of CarveNet. In Fig.~\ref{fig:dropmix}, we illustrate SensorAug for producing the partial point clouds. We denote a complete point cloud as $\mathcal{G}$. We construct a restricted 3D space centering at $\mathcal{G}$, where the coordinates of all of the points in $\mathcal{G}$ are normalized to the range [-0.5,0.5]. We randomly put a virtual LiDAR sensor, whose distance to the center of $\mathcal{G}$ is 1, to observe the object. The viewing frustum of the sensor is set to the default in \cite{shapenet2015}. The visible points within the viewing frustum is used for constructing the partial point cloud. We repeat SensorAug to produce a set of partial point clouds (\eg, $\{P1, P2, P3\}$ in Fig.~\ref{fig:dropmix}) for each complete point cloud.

For each complete point cloud $\mathcal{G}$, we use SensorAug to generate $T$ partial point clouds $\{\mathcal{P}_1$,...,$\mathcal{P}_T\}$. These generated point clouds are used, along with $\mathcal{P}$ given in the training set, by CarveNet to compute the coarse completion results $\{\mathcal{C},\mathcal{C}_1,...,\mathcal{C}_T\}$ and the refined results $\{\mathcal{Q},\mathcal{Q}_1,...,\mathcal{Q}_T\}$. We define the loss function below to optimize CarveNet, as:
\begin{align}
    \label{eq:floss}
    \mathcal{L} = \mathcal{L}_\mathrm{comp} + \alpha\mathcal{L}_\mathrm{sim}.
\end{align}
We minimize the loss function $\mathcal{L}$ to optimize CarveNet.

The loss $\mathcal{L}_\mathrm{comp}$ is formulated as:
\begin{align}
    \label{eq:loss_comp}
    \mathcal{L}_\mathrm{comp} &= \text{CD}(\mathcal{C}, \mathcal{G}) + \text{CD}(\mathcal{Q}, \mathcal{G}),\\
    \label{eq:cd}
    \text{CD}(\mathcal{X}_1, \mathcal{X}_2) &=  \frac{1}{|\mathcal{X}_1|}\sum_{\mathbf{p}_i\in\mathcal{X}_1}\min(\{\|\mathbf{p}_i-\mathbf{p}_j\|_2^2~|~\mathbf{p}_j\in\mathcal{X}_2\}) \nonumber\\
    & + \frac{1}{|\mathcal{X}_2|}\sum_{\mathbf{p}_i\in\mathcal{X}_2}\min(\{\|\mathbf{p}_i-\mathbf{p}_j\|_2^2~|~\mathbf{p}_j\in\mathcal{X}_1\}).
\end{align}
In Eq.~\eqref{eq:cd}, $\text{CD}$ denotes the Chamfer distance \cite{fan2017point} between a pair of point clouds $(\mathcal{X}_1, \mathcal{X}_2)$. We use the loss $\mathcal{L}_\mathrm{comp}$ to penalize the difference between the coarse/refined completion result and the ground-truth point cloud $\mathcal{G}$.

In Eq.~\eqref{eq:floss}, the loss $\mathcal{L}_\mathrm{sim}$ is defined as:
\begin{align}\label{eq:lsim}
    \mathcal{L}_\mathrm{sim} = \sum_{i=1}^T \text{CD}(\mathcal{C}_{i}, \mathcal{C}) +\text{CD}(\mathcal{Q}_i, \mathcal{Q}).
\end{align}
We use the loss $\mathcal{L}_\mathrm{sim}$ to penalize the difference between the completion results, which are computed based on different partial point clouds of the same object.

\noindent{\bf Qualitative discussion.}
In Fig.~\ref{fig:carvenet}(b), we provide four examples of the cell-wise convolution on different cells in the grid $\hat{\mathcal{B}}$. We train CarveNet on the ShapeNet dataset~\cite{shapenet2015}. We compare the partial point cloud $\mathcal{P}$ and the coarse grid $\hat{\mathcal{C}}$, where the regions of the 3D space are indicated by the circle and rectangle, respectively. CarveNet reasonably completes the missing grids (see the blue circle and rectangle), removes the grids beyond the object (see the yellow circle and rectangle), and preserves the correct details of the partial point cloud (see the red and pink circles and rectangles).

Yet, there is still much room for improving the robustness of CarveNet. In Fig.~\ref{fig:wodropmix}, given the different patterns of the partial point clouds of the identical object, CarveNet produces the inconsistent completion results. One of the major reasons, which lead to the inconsistent results, is the insufficient learning from the the relationship between the partial and the complete point clouds. This motivates us to propose SensorAug, which produces an arbitrary number of the partial point clouds for the same object. These partial point clouds are provided with diverse patterns. With SensorAug, we can use more pairs of the partial and the complete point clouds, for augmenting the training data and improving the completion power of CarveNet.

\subsection{Implementation Details}
\label{subsec:impl}

We use PyTorch 1.6 to implement CarveNet. We use Adam solver for network optimization. The mini-batch size is set to 24/32, for CarveNet with/without SensorAug. We use four NVIDIA 2080Ti GPUs for training and testing. The initial learning rate is set to $1e-4$. It is decayed linearly by a half for every 40 epoches. We set $\alpha=0.5$ and $T=2$.

%% file: S5_experiment.tex
\section{Experiments}\label{sec:experiments}

\subsection{Experimental Settings}\label{subsec:setups}

\noindent{\bf Dataset.} We use the ShapeNet \cite{shapenet2015} and KITTI \cite{Geiger2013IJRR} datasets to evaluate the completion methods. The ShapeNet dataset consists of 30,974 3D models from 8 categories. There are 28,974/800/1,200 models in the training/validation/test set. Each model is associated with the a pair of complete and partial point clouds. The complete point cloud contains 16,384 points uniformly sampled on the object surface, while the partial one contains 2,048 points.
The point clouds in KITTI \cite{Geiger2013IJRR} are captured by the LiDAR sensors. There are 2,400 partial point clouds of cars, which are taken from 426 different timestamps. Each cloud contains 2,048 points. The ground-truth complete clouds are unavailable in KITTI.

\noindent{\bf Baselines.} We compare our method with FoldingNet \cite{yang2018foldingnet}, PCN \cite{yuan2018pcn}, AltasNet \cite{groueix2018papier}, TopNet \cite{tchapmi2019topnet}, and GRNet \cite{xie2020grnet}. All of these methods are compared fairly with the same experimental settings. Here, we set the number of patches to 16 in AtlasNet \cite{groueix2018papier}. The number of levels and leaves were set to 6 and 8 in TopNet \cite{tchapmi2019topnet} for generating 16,384 points.

\begin{table}[b]
\scriptsize
\centering
\vspace{-10pt}
\begin{tabular}{l| c c }
\toprule
Method & Coarse CD & Dense CD  \\
\midrule
w/o Construction & 7.218 & 0.5072\\
Symmetric ($2048$ pts) & 1.929 & 0.4624\\
\hline
Uniform ($11^3$ pts) & \textbf{1.754} & 0.4062\\
Uniform ($13^3$ pts) & 1.782 & \textbf{0.3905}\\
Uniform ($16^3$ pts) & 2.223 & 0.3907\\
\hline
Ground Truth Pts & 1.561 & 0.3722\\
\bottomrule
\end{tabular}
\caption{The comparison of using different initial points to construct the point-block. Coarse/Dense CD represent the average Chamfer distance (multiplied by $10^3$) between the coarse/refine results and ground-truth point clouds.}\vspace{-10pt}
\label{tab:ab1}
\end{table}
%

\noindent{\bf Evaluation metrics.}
We use Chamfer distance and Consistency as the metrics for the evaluation. We use Chamfer distance (see Eq.~\eqref{eq:cd}) to measure the similarity between completion result $\mathcal{Q}$ and the complete point cloud $\mathcal{G}$. Because the ground-truth point clouds are unavailable in the KITTI dataset, we use the consistency defined in \cite{yuan2018pcn, xie2020grnet} to measure the quality of the completion result. We denote the completion result of $i^{th}$ car in $j^{th}$ frame as $\mathcal{Q}^i_{j}$. Suppose there are $N$ frames for the $i^{th}$ car. The completion results of the adjacent frames are used for computing the Chamfer distances, which are averaged to achieve the consistency as:
\begin{align}\label{consistency}
\text{Consistency} = \frac{1}{N-1}\sum_{j=1}^{N-1} \text{CD}(\mathcal{Q}^i_{j}, \mathcal{Q}^i_{j+1})
\end{align}
A lower Chamfer distance/consistency means a better result.

\subsection{Ablation Study}\label{sec:ablation}

We use the ShapeNet dataset for evaluating the core components of CarveNet, \ie, the point-block construction, the 3D point cloud representation and SensorAug. We disable SensorAug for examining the improvements, which are solely contributed by the point-block construction and the 3D point cloud representation, respectively.

\begin{figure}[t]
\centering
\includegraphics[width=0.9\linewidth]{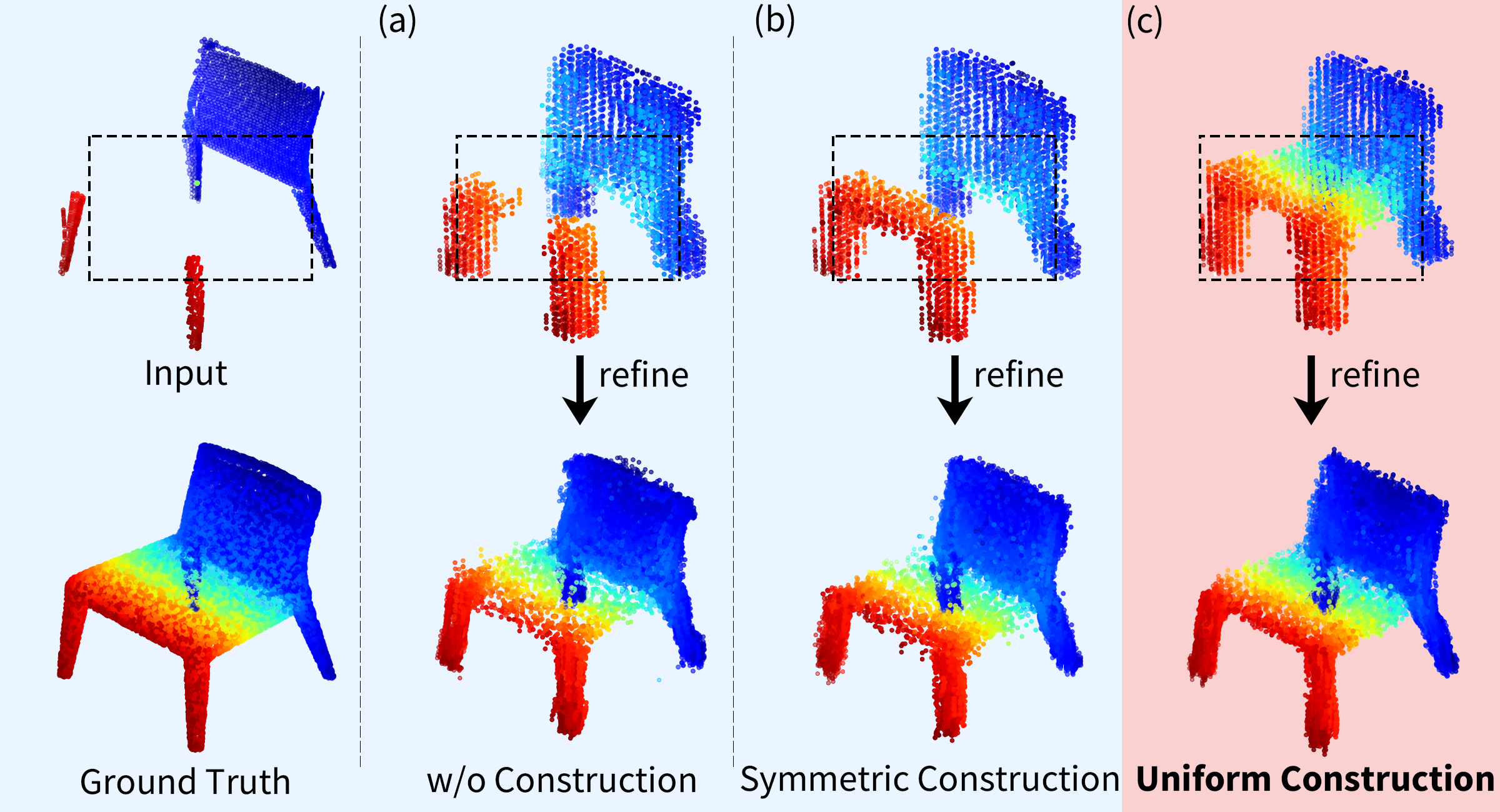}
\caption{The completion results achieved by different point-block construction methods. The first column shows the input and the ground truth. The first and second rows of other columns are coarse completion results (\ie, after point-block engraving) and dense completion results (\ie, after refinement) of each method, respectively.}
\label{fig:ab1}
\end{figure}


\begin{figure*}[t]
\centering
\includegraphics[width=0.9\linewidth]{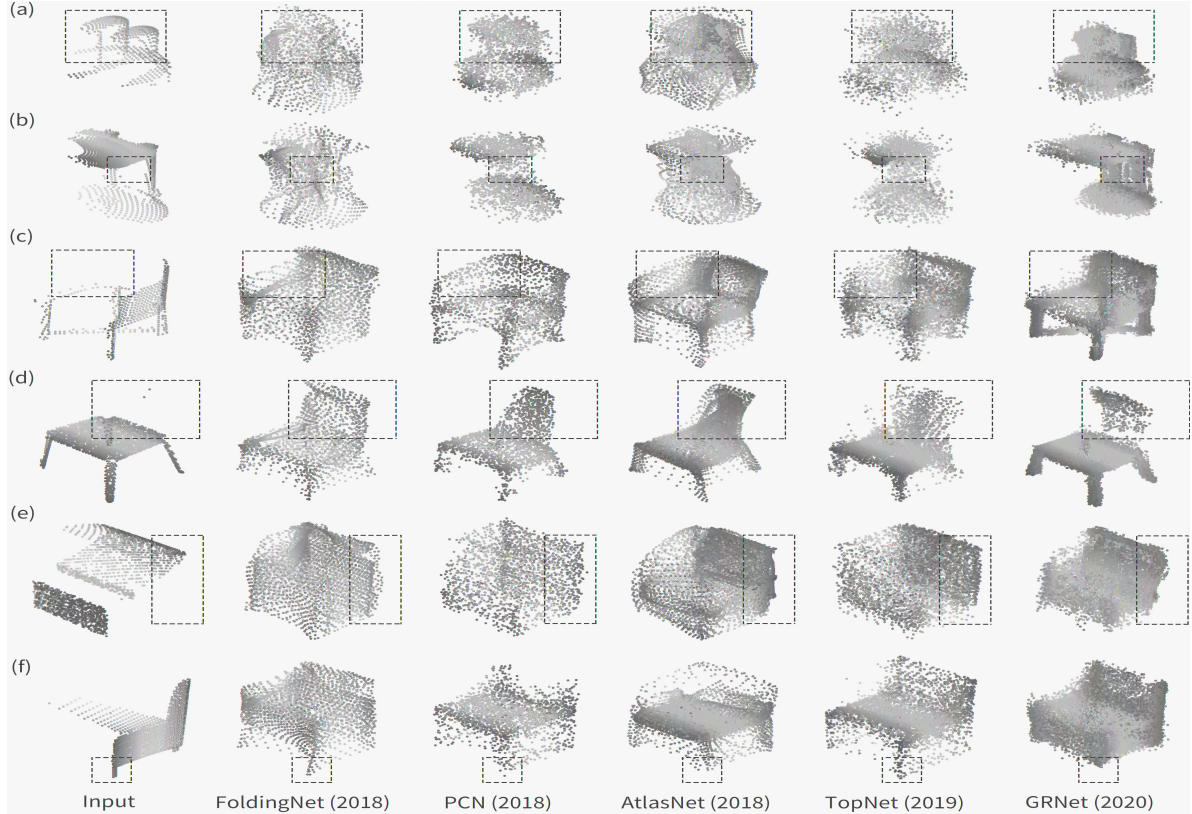}
   \caption{Visualization of the point cloud completion results of different methods on the ShapeNet dataset.}
\label{fig:eva1}
\vspace{-0pt}
\end{figure*}

\noindent{\bf Evaluation of the point-block construction methods.}
In Table~\ref{tab:ab1}, we report the average Chamfer distances for all of the completion results on the ShapeNet test set. In the row \textit{w/o Construction}, we show the results of carving on the point-block, which contains the partial point cloud only. Because many objects have symmetric shapes, a naive solution for the point-block construction is simply merging the partial point clouds and its mirror counterpart as an entire point-block (see the results in the row \textit{Symmetric}). For our method (\ie, uniform sampling), we compare different number of points (\ie, $11^3, 13^3$ and $16^3$ points).

In the column \textit{Coarse CD}, we compare the rdifferent point-block construction methods in terms of the qualities of the coarse completion results. Our strategy of the uniform point sampling outperforms other alternatives for the point-block construction. The completion results of the compared methods are visualized in Fig.~\ref{fig:ab1}. By comparing Fig.~\ref{fig:ab1} (a--c), we find that the point-block construction helps the point-block engraving to better recover the 3D points lost in the partial point cloud. We also evaluate our point cloud refinement, which is used along with different point-block construction methods. The refinement produces the dense completion results, whose qualities are reported in the column \textit{Dense CD}. Our refinement consistently improves the completion qualities achieved by different methods.

In Table~\ref{tab:ab1}, we use different numbers of the sampled points to construct the point-block and compare the completion qualities. Though more sampled points (\eg, $16^3$ points) may provide more chances for recovering the object details, they significantly increase the difficulty of carving and lead to worse completion. Here, we investigate an extreme strategy, where the ground-truth cloud is given for constructing the point-block. Compared to the extreme strategy, using the uniform points produces the competitive results. Because the prior object information is unnecessary, the uniform point sampling for constructing the point-block can be generalized to the completion of objects in different categories.

\begin{table}[b]
\scriptsize
\centering
\vspace{-10pt}
    \begin{subtable}[t]{0.48\linewidth}
    \centering
        \begin{tabular}{l|c c c }
            \toprule
            Method & CD  \\
            \midrule
            KPConv \cite{thomas2019kpconv}         & 0.6628 \\
            PointConv \cite{wu2019pointconv}       & 0.6448 \\
            3D Grids & \textbf{0.3905} \\
            \bottomrule
        \end{tabular}
        \caption{The effect of using different 3D point representations.}
    \end{subtable}
    \begin{subtable}[t]{0.48\linewidth}
    \centering
        \begin{tabular}{l|c c c }
            \toprule
            Method & CD  \\
            \midrule
            w/o SensorAug                & 0.3905 \\
            Randomly Drop       & 0.4463 \\
            {SensorAug}  & \textbf{0.3833} \\
            \bottomrule
        \end{tabular}
        \caption{The effect of CarveNet with/without SensorAug.}
    \end{subtable}
\vspace{5pt}
\caption{Ablation study on the 3D representation and SensorAug. Chamfer distance (CD) is multiplied by $10^3$.}\vspace{-10pt}
\label{tab:ab}
\end{table}

\noindent{\bf Evaluation of the 3D point cloud representations.}
In our implementation of CarveNet, We use the 3D grid-based intermediate representation for the point-block engraving. In Table~\ref{tab:running time}(a), we compare the 3D grid-based and different point-based representations and their performances on the ShapeNet test set. Note that the point-based representation disables the conventional operation of the 3D convolution. Thus, we resort to KPConv \cite{thomas2019kpconv} and PointConv \cite{wu2019pointconv} for the comparison with the 3D point-based representations. The 3D grid-based method produces better result than the point-based representations. In Table~\ref{tab:running time}, we compare the 3D grid-based method with different grid sizes. The running time is measured as the forwarding time with a batch size of 1. By considering the completion accuracy and the computation, we select the grid size $64^3$in our implementation.

%
\begin{table}[t]
\scriptsize
\centering
\vspace{-5pt}
\begin{tabular}{l| c c c}
\toprule
Grid Size & CD ($\times 10^{3}$) & Running time (ms) & \# Parameters (M)  \\
\midrule
$32^3$ & 0.6486 & 15 & 6.9 \\
$64^3$ & 0.3905 & 29 & 76.8\\
$80^3$ & 0.4010 & 77 & 178.7\\
\bottomrule
\end{tabular}
\caption{\scriptsize Performance on the ShapeNet test set with different sizes of inputs.}\vspace{-5pt}
\label{tab:running time}
\end{table}

\begin{table*}[tbp]
\scriptsize
\centering
\begin{tabular}{l|c c c c c c c c |c}
\toprule
Method & Airplane & Cabinet & Car & Chair & Lamp & Sofa & Table & Vessel & Average  \\
\midrule
FoldingNet \cite{yang2018foldingnet} & 0.4127 & 0.6837 & 0.4096 & 0.8226 & 0.8475 & 0.7212 & 0.6478 & 0.4778 & 0.6279 \\
PCN \cite{yuan2018pcn} & 0.3431 & 1.0998 & 0.5513 & 1.0955 & 1.1840 & 1.2163 & 1.0459 & 0.6917 & 0.9035 \\
AtlasNet \cite{groueix2018papier} & 0.2503 & 0.6860 & 0.3832 & 0.6912 & 0.8178 & 0.8135 & 0.5971 & 0.5169 & 0.5945 \\
TopNet \cite{tchapmi2019topnet} & \textbf{0.1711} & 0.5319 & 0.3565 & 0.5947 & 0.5518 & 0.6735 & 0.3935 & 0.3710 & 0.4555\\
GRNet \cite{xie2020grnet} & 0.2840 & 0.5966	& 0.3347 & 0.5353 & 0.4486 & 0.7456 & 0.5066 & 0.3011 & 0.4691 \\
\textbf{CarveNet (Ours)} & 0.2043 & \textbf{0.4988} & \textbf{0.3005} & \textbf{0.4584} & \textbf{0.3976} & \textbf{0.5469} & \textbf{0.3737} & \textbf{0.2862} & \textbf{0.3833}\\
\bottomrule
\end{tabular}
\caption{The results on the ShapeNet. Here, we compute Chamfer distance based on 16,384 points (multiplied by $10^3$). The best result of each column is bold-face. }\vspace{0pt}
\label{tab:eva1}
\end{table*}
%

\begin{figure}[t]
\centering
\includegraphics[width=0.95\linewidth]{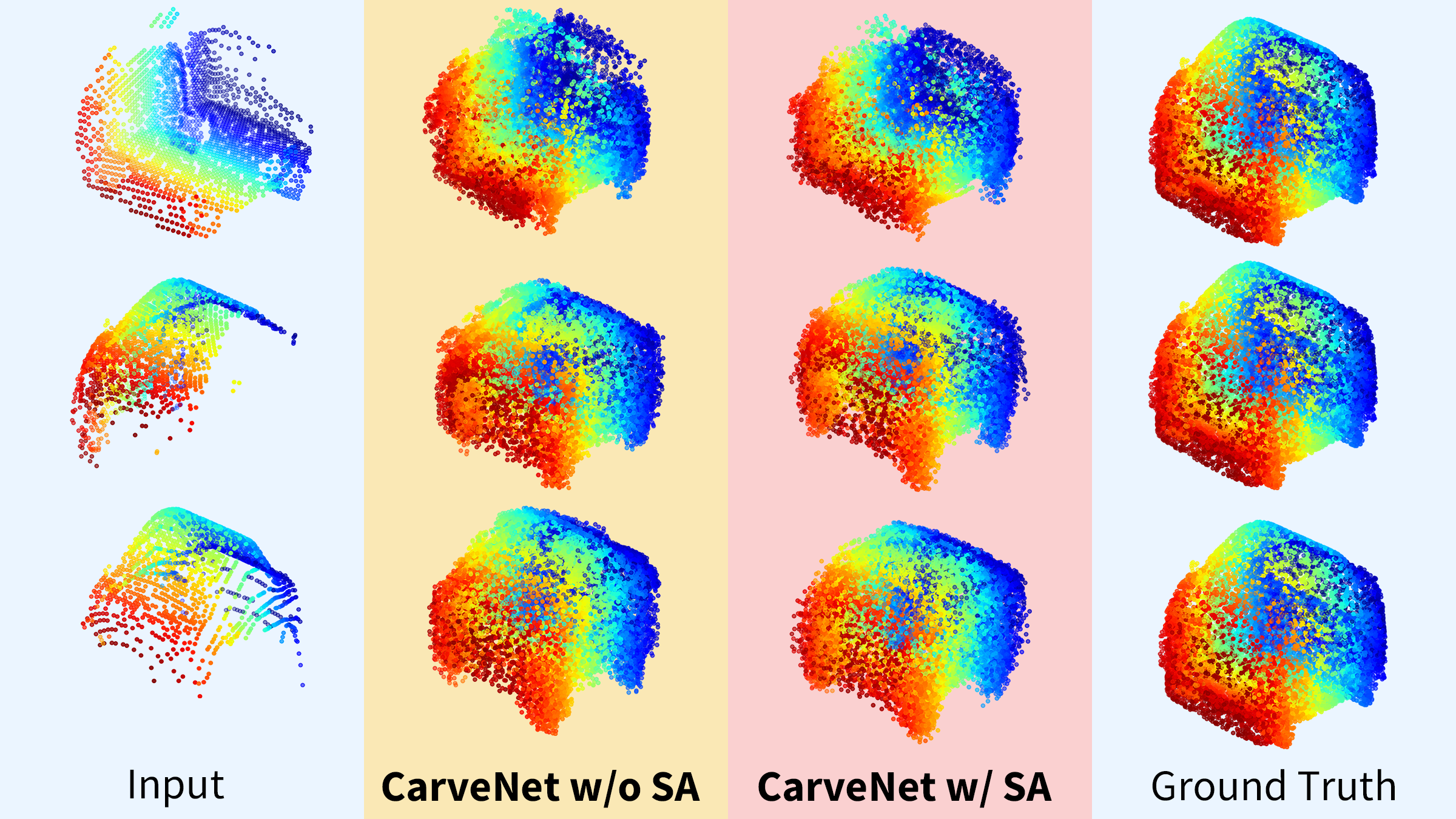}
   \caption{Completion results without/with SensorAug for training. Here, different partial point clouds capture the same object.}
\label{fig:dropmix_comp}
\end{figure}

\noindent{\bf Evaluation of SensorAug.}
We examine the effect of removing SensorAug from the network training and report the completion result in Table~\ref{tab:ab} (b). Without SensorAug (see the row \textit{w/o SensorAug}), we successfully degrade the completion results. We also compare SensorAug with the random dropout of points (see the row \textit{Randomly Drop}) for augmenting the training data. The random dropout involves inconsistent object shapes that mislead the network training, thus yielding lower completion accuracy than the method without augmentation and SensorAug. We visualize the completion results without/with SensorAug in Fig.~\ref{fig:dropmix_comp}. Given different partial point clouds, the network, which is trained with SensorAug, produces consistent completion results.

\begin{table}[]
\scriptsize
\centering
\begin{tabular}{l|c c c }
\toprule
Method & Consistency ($\times10^3$) \\
\midrule
FoldingNet \cite{yang2018foldingnet}& 0.2854 \\
PCN \cite{yuan2018pcn}              & 0.3578 \\
AtlasNet \cite{groueix2018papier}   & 0.3580 \\
TopNet \cite{tchapmi2019topnet}     & 0.2179 \\
GRNet \cite{xie2020grnet}           & 0.2036 \\
\textbf{CarveNet}                   & \textbf{0.1745} \\
\bottomrule
\end{tabular}
\caption{Consistency on KITTI.}
\label{tab:kitti_dropmix_exp}
\end{table}

\subsection{Comparison with State-of-the-Art Methods}

\noindent{\bf Results on ShapeNet.}
In Table~\ref{tab:eva1}, we compare CarveNet with other methods. Here, we evaluate each method in terms of the average Chamfer distances on 8 object categories, respectively. The distances on 8 categories are averaged again, for measuring the overall performance of the method. Our method surpasses other methods on 7 of 8 categories. We also provide the examples of the completion results in Fig.~\ref{fig:eva1}, where the complex shapes like high-curvature and hollowed-out object parts are recovered properly.

\begin{figure}[t]
\centering
\vspace{-0pt}
\includegraphics[width=\linewidth]{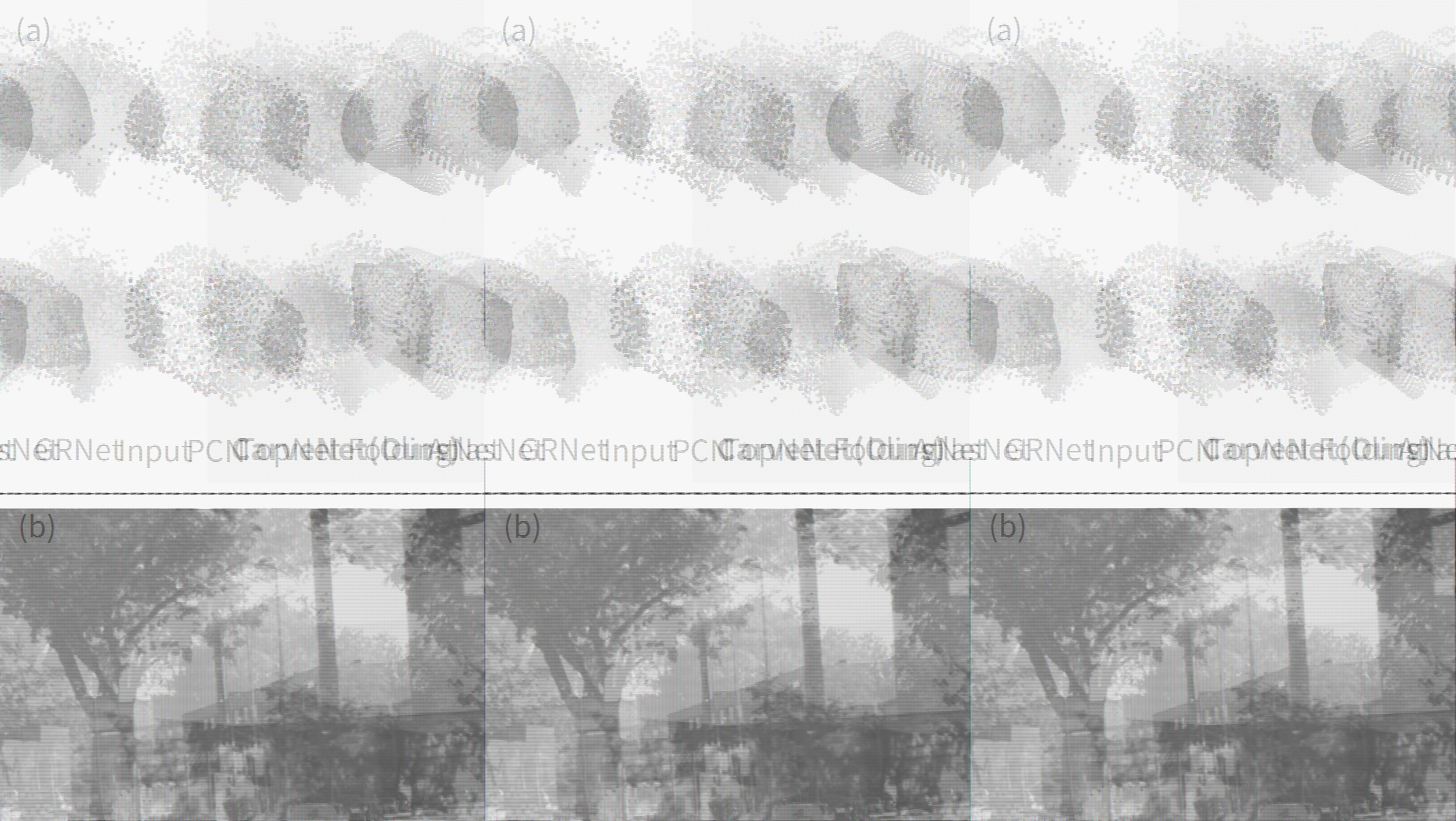}
\caption{\scriptsize Visualization of completion results on KITTI. (a) Completion results through diffrent methods. (b) Completion results of our method projecting on image. Top: partial point clouds; Bottom: completed results.}
\label{fig:kitti}
\vspace{-0pt}
\end{figure}

\noindent{\bf Results on KITTI.}
Because the ground-truth point clouds are unavailable, we train CarveNet on the car models in the ShapeNet training set. The completion results achieved by CarveNet are evaluated on the KITTI test set. In Table~\ref{tab:kitti_dropmix_exp}, we report the consistencies achieved by different methods. Again, CarveNet outperforms other methods. It also demonstrates that knowledge learned by CarveNet can be generalized to different datasets. We provide the examples of the completion results on the KITTI dataset in Fig.~\ref{fig:kitti}.

\begin{figure}[t]
\centering
\vspace{-0pt}
\includegraphics[width=0.95\linewidth]{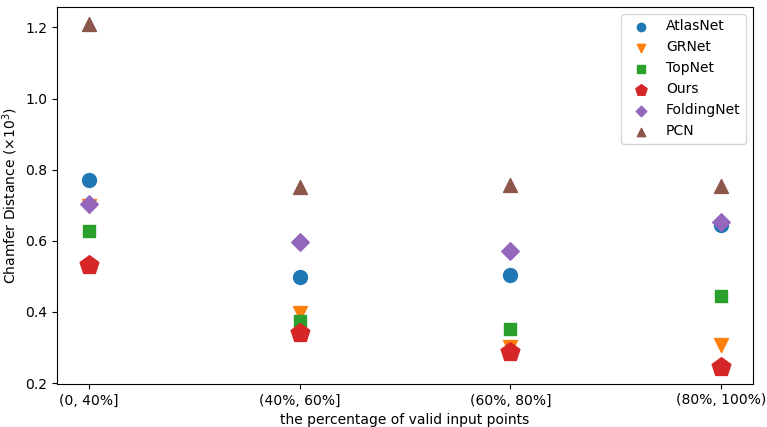}
   \caption{Sensitivity to the percentage of valid points in the partial point clouds.}
\label{fig:robust}\vspace{-0pt}
\end{figure}

\noindent{\bf Sensitivity to the partial point clouds.}
In the default setting, each partial point cloud contains 2,048 points. During the completion, we evaluate the sensitivities of different methods to the number of valid points in the partial point clouds. We reduce the percentage of the valid points in the partial point clouds. Note that calculating the percentage of the valid points directly cannot well-illustrate the missing of structural information in the partial point cloud comparing to the ground truth, thus, here we group points into $64\times64\times64$ grids, and define the percentage of valid points as the percentage of non-zero grids of input compared to ground truth. The valid points are then fed to the trained model for completion. We compare the completion accuracies of different methods in Fig.~\ref{fig:robust}. With different percentages of the valid points, CarveNet produces better results than other methods. 

%% file: S6_conclusion.tex
\section{Conclusion}\label{sec:concl}

The complexity of 3D point cloud completion stems from the the diversity of the 3D object shapes. In this paper, we have proposed an effective operation, the point-block engraving, for the completion task. We use the point-block engraving to manipulate on the the 3D grid-based representation of the object, which is shape-agnostic, for removing the redundant points and recovering the important details of the object. Moreover, we propose SensorAug to augment the training data, allowing the completion network to learn from more diverse object shapes. The evaluation on the public completion benchmarks demonstrates the effectiveness of our approach.
